\documentclass[letterpaper]{article} 
\usepackage{aaai24}  
\usepackage{times}  
\usepackage{helvet}  
\usepackage{courier}  
\usepackage[hyphens]{url}  
\usepackage{graphicx} 
\usepackage{amssymb}
\usepackage{amsmath}
\usepackage{multirow}
\usepackage{rotating}
\usepackage{utfsym}
\usepackage{float}
\usepackage{subfig}
\usepackage{booktabs}
\usepackage{natbib}  
\usepackage{caption} 
\usepackage{algorithm}
\usepackage{algorithmic}

\usepackage{newfloat}
\usepackage{listings}
\DeclareCaptionStyle{ruled}{labelfont=normalfont,labelsep=colon,strut=off} 
\floatstyle{ruled}
\newfloat{listing}{tb}{lst}{}
\floatname{listing}{Listing}
\title{FontDiffuser: One-Shot Font Generation via Denoising Diffusion with \\Multi-Scale Content Aggregation and Style Contrastive Learning}
\author{
    Zhenhua Yang\textsuperscript{\rm 1},
    Dezhi Peng\textsuperscript{\rm 1},
    Yuxin Kong\textsuperscript{\rm 1},
    Yuyi Zhang\textsuperscript{\rm 1},
    Cong Yao\textsuperscript{\rm 3},
    Lianwen Jin\textsuperscript{\rm 1\rm2}\thanks{Corresponding author}
}
\affiliations{
    \textsuperscript{\rm 1}South China University of Technology, 
    \textsuperscript{\rm 2}SCUT-Zhuhai Institute of Modern Industrial Innovation, 
    \textsuperscript{\rm 3}Alibaba Group\\

    \{eezhyang, pengdezhi000, kongyxscut, yuyizhang.scut, yaocong2010\}@gmail.com, 
    eelwjin@scut.edu.cn
}

\usepackage{bibentry}

\begin{document}

\maketitle

\begin{abstract}
Automatic font generation is an \textit{imitation task}, which aims to create a font library that mimics the style of reference images while preserving the content from source images. 
Although existing font generation methods have achieved satisfactory performance, they still struggle with \textit{complex characters} and \textit{large style variations}. 
To address these issues, we propose \textbf{FontDiffuser}, a diffusion-based image-to-image one-shot font generation method, which innovatively models the font imitation task as a noise-to-denoise paradigm. 
In our method, we introduce a Multi-scale Content Aggregation (MCA) block, which effectively combines global and local content cues across different scales, leading to enhanced preservation of intricate strokes of complex characters. 
Moreover, to better manage the large variations in style transfer, we propose a Style Contrastive Refinement (SCR) module, which is a novel structure for style representation learning. 
It utilizes a style extractor to disentangle styles from images, subsequently supervising the diffusion model via a meticulously designed style contrastive loss. 
Extensive experiments demonstrate FontDiffuser's state-of-the-art performance in generating diverse characters and styles. It consistently excels on complex characters and large style changes compared to previous methods. 
The code is available at https://github.com/yeungchenwa/FontDiffuser.
\end{abstract}

\section{Introduction}
\noindent Automatic font generation aims to create a new font library in the required style given the reference images, which is referred to as an \textit{imitation task}. 
Font generation has significant applications, including new font creation, ancient character restoration, and data augmentation for optical character recognition. 
Therefore, it has significant commercial and cultural values. However, this imitation process is both costly and labor-intensive, particularly for languages with a large number of glyphs, such as Chinese ($>$ 90,000), Japanese ($>$ 50,000), and Korean ($>$ 11000). Existing automatic methods primarily disentangle the representations of style and content, then integrate them to output the results.

\begin{figure}[t]
    \centering
    \subfloat[Characters generated by our method]{
        \includegraphics[width=0.98\columnwidth]{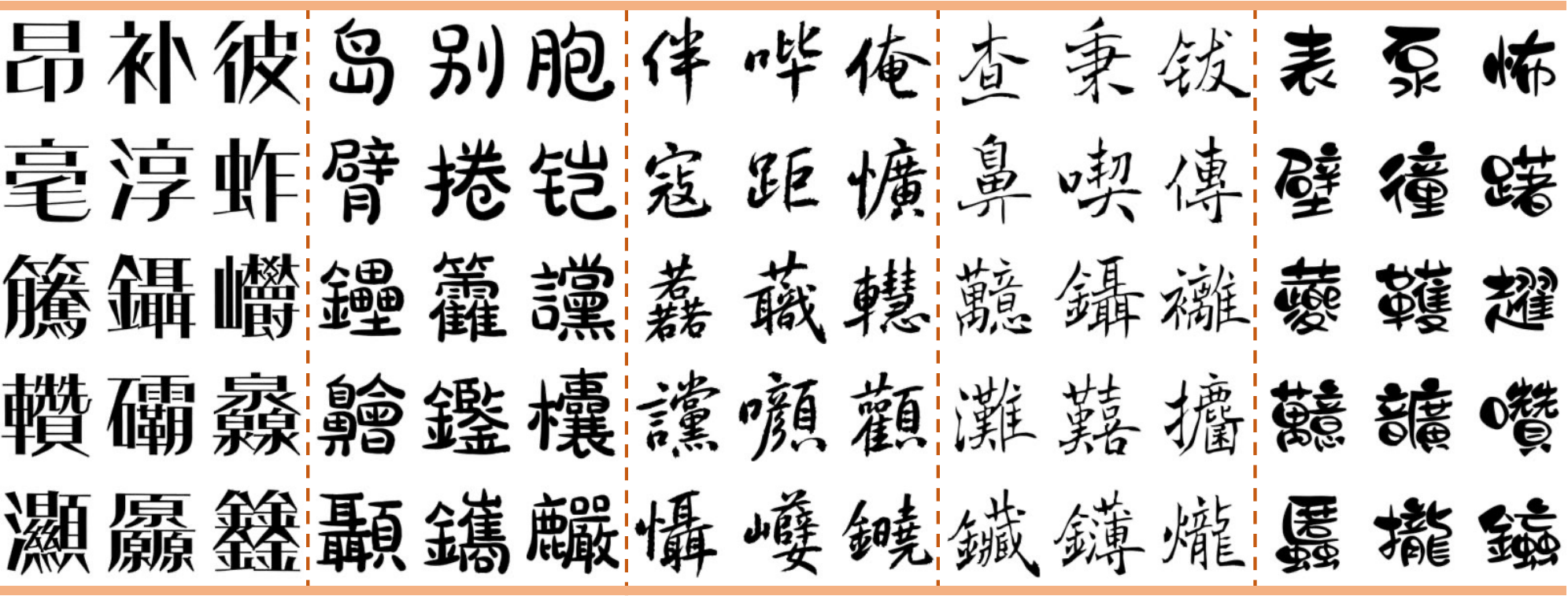}}
    \quad
    \subfloat[Complex characters]{
        \includegraphics[width=0.47\columnwidth]{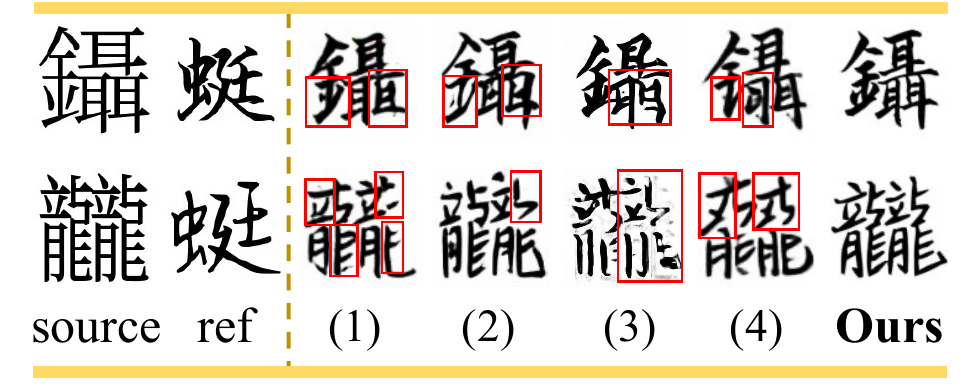}}
    \subfloat[Large style variations]{
        \includegraphics[width=0.47\columnwidth]{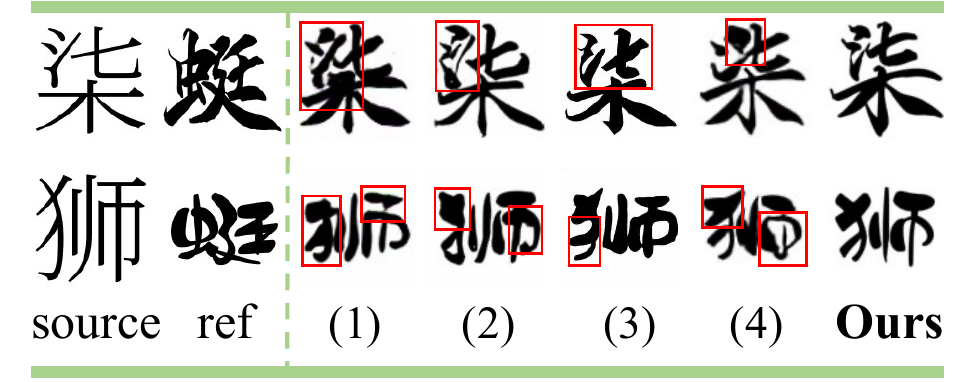}}
    \caption{(a) Characters of different complexity generated by our method. (b)(c) Results of different methods on complex characters and large style variations. `ref' represents the reference image. (1)-(4) represent the results of DG-Font \cite{xie2021dg}, MX-Font \cite{park2021multiple}, CG-GAN \cite{kong2022look}, and CF-Font \cite{wang2023cf} respectively. Red boxes highlight the failures of other methods.}
    \label{fig:generated}
\end{figure}

Although these methods have achieved remarkable success in font generation, they still suffer from \textit{complex character generation} and \textit{large style variation transfer}, leading to severe stroke missing, artifacts, blurriness, layout errors, and style inconsistency as shown in Figure \ref{fig:generated}(b)(c).
Retrospectively, most font generation approaches \cite{park2021few, park2021multiple, xie2021dg, tang2022few, liu2022xmp, kong2022look, wang2023cf} adopt a GAN-based~\cite{goodfellow2014generative} framework which potentially suffers from unstable training due to their adversarial training nature. 
Moreover, most of these methods perceive content information through only single-scale high-level features, omitting the fine-grained details that are crucial to preserving the source content, especially for complex characters.
There are also a number of methods \cite{cha2020few, park2021few, park2021multiple, liu2022xmp, kong2022look, he2022diff} that employ prior knowledge to facilitate font generation, such as stroke or component composition of characters; however, this information is costly to annotate for complex characters.
Furthermore, the target style is commonly represented by a simple classifier or a discriminator in previous literature, which struggles to learn the appropriate style and hinders the style transfer with large variations.

In this paper, we propose FontDiffuser, a diffusion-based image-to-image one-shot font generation method, which models the font generation learning as a noise-to-denoise paradigm and is capable to generate unseen characters and styles.
In our method, we innovatively introduce a Multi-scale Content Aggregation (MCA) block, which leverages global and local content features across various scales. This block effectively preserves intricate details from the source image of complex characters, by capitalizing on the fact that large-scale features contain lots of fine-grained information (strokes or components), whereas small-scale features primarily encapsulate global information (layout).
Moreover, we introduce a novel style representation learning strategy, by applying a Style Contrastive Refinement (SCR) module to enhance the generator's capability in mimicking styles, especially for large variations between the source image and the reference image. This module utilizes a style extractor to disentangle style from a font and then uses a style contrastive loss to provide feedback to the diffusion model. SCR acts as a supervisor and encourages our diffusion model to identify the differences among various samples, which are with different styles but the same character. Additionally, we design a Reference-Structure Interaction (RSI) block to explicitly learn structural deformations (\textit{e.g.}, font size) by utilizing a cross-attention interaction with the reference features. 

To verify the effectiveness of generating characters of diverse complexity, we categorize the characters into three levels of complexity (easy, medium, and hard) according to their number of strokes, and test our method on each level separately. Extensive experiments demonstrate that our proposed FontDiffuser outperforms state-of-the-art font generation methods on characters of three levels of complexity. Notably, as shown in Figure \ref{fig:generated}(a), FontDiffuser consistently excels both in the generation of complex characters and large style variations. Furthermore, our method can be applied to the cross-lingual generation tasks, showcasing the cross-domain generalization ability of FontDiffuser. 

We summarize our main contributions as follows.
\begin{itemize}
\item[$\bullet$] 
We propose FontDiffuser, a new diffusion-based image-to-image one-shot font generation framework that achieves state-of-the-art performance in generating complex characters and handling large style variations.
\item[$\bullet$] 
To enhance the preservation of intricate strokes of complex characters, we propose a Multi-scale Content Aggregation (MCA) block, leveraging the global and local features across different scales from the content encoder. 
\item[$\bullet$] 
We propose a novel style representation learning strategy and elaborate a Style Contrastive Refinement (SCR) module that supervises the diffusion model using a style contrastive loss, enabling effective handling of large style variations.
\item[$\bullet$] 
FontDiffuser demonstrates superior performance over existing methods in generating characters across easy, medium, and hard complexity levels, showcasing strong generalization capability across unseen characters and styles. Furthermore, our method can be extended to the cross-lingual generation, such as Chinese to Korean.
\end{itemize} 

\section{Related Work}
\subsection{Image-to-image Translation} 
Image-to-Image (I2I) translation task is to convert an image from a source domain into a target domain. Previously, image-to-image methods \cite{isola2017image, liu2017unsupervised, zhu2017unpaired, liu2019few} are commonly tackled through GAN \cite{goodfellow2014generative}. For instance, Pix2pix \cite{isola2017image} is the first I2I translation framework. FUNIT \cite{liu2019few} utilizes AdaIN \cite{huang2017arbitrary} to combine the encoded content image and class image. Recently, there have been numerous methods \cite{choi2021ilvr, sasaki2021unit, saharia2022palette} utilizing diffusion models to address image-to-image translation tasks. For example, ILVR \cite{choi2021ilvr} generates high-quality images based solely on a trained DDPM \cite{ho2020denoising} using a reference image. Palette \cite{saharia2022palette} proposes a simple image-to-image diffusion model and outperforms GAN and regression baselines.

\subsection{Few-shot font generation} Early font generation methods \cite{chang2018chinese, lyu2017auto, zi2zi, jiang2017dcfont, sun2018pyramid} consider the font generation task as an image-to-image translation problem, but they cannot generate unseen style fonts. To address this, SA-VAE \cite{sun2017learning} and EMD \cite{zhang2018separating} generate unseen fonts by disentangling style and content representations. To enable the generator to capture local style characteristics, some methods \cite{wu2020calligan, huang2020rd, cha2020few, park2021few, park2021multiple, liu2022xmp, kong2022look} utilize prior knowledge, such as stroke and component. For instance, LF-Font \cite{park2021few}, MX-Font \cite{park2021multiple} and CG-GAN \cite{kong2022look} employ a component-based learning strategy to enhance the capability of local style representation learning. XMP-Font \cite{liu2022xmp} utilizes a pre-training strategy to facilitate the disentanglement of style and content. Diff-Font \cite{he2022diff} adopts stroke information to support the sampling but fails to generate unseen characters. However, the annotation of strokes and components is costly for complex characters. Some prior-free methods \cite{xie2021dg, tang2022few, wang2023cf} have been proposed. DG-Font \cite{xie2021dg} achieves promising performance in an unsupervised manner. Fs-Font \cite{tang2022few} aims to discover the spatial correspondence between content images and style images to learn the local style details, but its reference selection strategy is sensitive to the quality of results. CF-Font \cite{wang2023cf} fuses various content features of different fonts and introduces an iterative style-vector refinement strategy. However, these methods still struggle with generating complex characters and handling large variations in style transfer. 

\begin{figure*}[t]
    \centering
    \includegraphics[width=\textwidth]{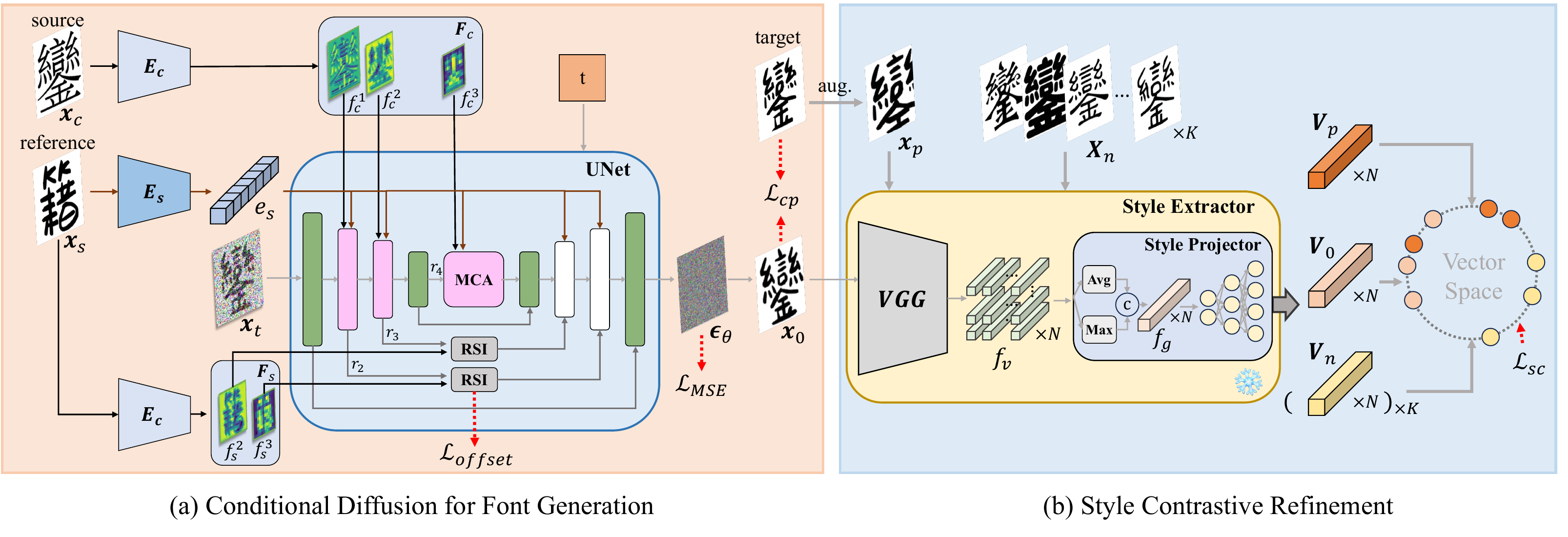}
    \caption{Overview of our proposed method. (a) The Conditional Diffusion model is a UNet-based network composed of a content encoder $E_{c}$ and a style encoder $E_{s}$. The reference image $\boldsymbol{x}_{s}$ is passed through a style encoder $E_{s}$ and a content encoder $E_{c}$ respectively, obtaining a style embedding $e_{s}$ and structure maps $\boldsymbol{F}_{s}$. The source image is encoded by a content encoder $E_{c}$. To obtain multi-scale features $\boldsymbol{F}_{c}$, we derive output from the different layers of $E_{c}$  and inject each of them through our proposed MCA block. RSI block is employed to conduct spatial deformation from reference structural features $\boldsymbol{F}_{s}$. (b) The Style Contrastive Refinement module is to disentangle different styles from images and provide guidance to the diffusion model.}
    \label{fig:framework}
\end{figure*}

\subsection{Diffusion model} Recently, diffusion models have achieved rapid development in vision generation tasks. Several prominent conditional diffusion models have been developed \cite{nichol2021glide, ramesh2022hierarchical, saharia2022photorealistic, rombach2022high, zhang2023adding, ruiz2023dreambooth}. For example, LDM \cite{rombach2022high} proposes a cross-attention mechanism to incorporate the condition into the UNet and treats the diffusion process in the latent space. In text image generation, \cite{luhman2020diffusion, gui2023zero, nikolaidou2023wordstylist} apply diffusion models to generate handwritten characters and demonstrate their promising effects. CTIG-DM \cite{zhu2023conditional} devises image, text, and style as conditions and introduces four text image generation modes in a diffusion model. In contrast to general image generation, font generation requires distinct stroke details and intricate structural features at a fine-grained level. This motivates us to harness multi-scale content features and propose an innovative style contrastive learning strategy. 

\section{Methodology}
\noindent As shown in Figure \ref{fig:framework}, our proposed method consists of a Conditional Diffusion model and a Style Contrastive Refinement module. In the \textbf{Conditional Diffusion} model, given a source image $\boldsymbol{x}_{c}$ and a reference image $\boldsymbol{x}_{s}$, our goal is to train a conditional diffusion model where the final output image should not only have the same content as in $\boldsymbol{x}_{c}$, but should also be consistent with the reference style. \textbf{Style contrastive refinement} module aims to disentangle different styles from a group of images and offer guidance to the diffusion model via a style contrastive loss.

\subsection{Conditional Diffusion for Font Generation}
\noindent Based on DDPM \cite{ho2020denoising}, the general idea of our diffusion-based image-to-image font generation method is to design a forward process that incrementally adds noise to the target distributions $\boldsymbol{x}_{0}\sim q(\boldsymbol{x}_{0})$, while the denoising process involves learning the reverse mapping. The denoising process aims to transform a noise $\boldsymbol{x}_{T}\sim (0, \boldsymbol{I})$ to the target distribution in $T$ steps. 

Specifically, the forward process of FontDiffusers is a Markov chain and the noise adding process can be summarized as follows:
\begin{align}
    \boldsymbol{x}_{t} = \sqrt{\bar{\alpha}_{t}}\boldsymbol{x}_{0}+\sqrt{1-\bar{\alpha}_{t}}\boldsymbol{\epsilon},  
\end{align}
where $t\sim [0, T]$, $\boldsymbol{\epsilon}$ is the added Gaussian noise. $\alpha_{t}=1-\beta_{t}$, $\bar{\alpha}_{t} = \prod_{i=0}^{t} (1 - \beta_{i})$, $\beta_{i} \sim (0, 1) $ is a fixed hyper-parameter of variance.
During the reverse process, the reverse mapping can be approximated by a model to predict the noise $\boldsymbol{\epsilon}_{\theta} (\boldsymbol{x}_{t}, t, \boldsymbol{x}_{c}, \boldsymbol{x}_{s})$ and then obtain the $\boldsymbol{x}_{t-1}$ as follows:
\begin{small}
    \begin{align}
    \boldsymbol{x}_{t-1} = \frac{1}{\sqrt{\alpha_{t}}}(\boldsymbol{x}_{t} - \frac{1-\alpha_{t}}{\sqrt{1-\bar{\alpha}_{t}}}\boldsymbol{\epsilon}_{\theta} (\boldsymbol{x}_{t}, t, \boldsymbol{x}_{c}, \boldsymbol{x}_{s}))+\sigma_{t}\boldsymbol{z}, 
    \end{align} 
    \label{equ:reverse}
\end{small}
where $\sigma_{t}$ is the hyper-parameter and noise $\boldsymbol{z}\sim (0, \boldsymbol{I})$.

We predict the noise $\boldsymbol{\epsilon}_{\theta} (\boldsymbol{x}_{t}, t, \boldsymbol{x}_{c}, \boldsymbol{x}_{s})$ using our conditional diffusion model. Specifically, to enhance the preservation of complex characters, we employ a \textit{Multi-scale Content Aggregation} (MCA) block to inject the global and local content cues into the UNet of our model. Moreover, a \textit{Reference-Structure Interaction} (RSI) block is employed to facilitate structural deformation from the reference features.

\subsubsection{Multi-scale Content Aggregation (MCA)} 
\begin{figure}[ht]
    \centering
    \includegraphics[width=\columnwidth]{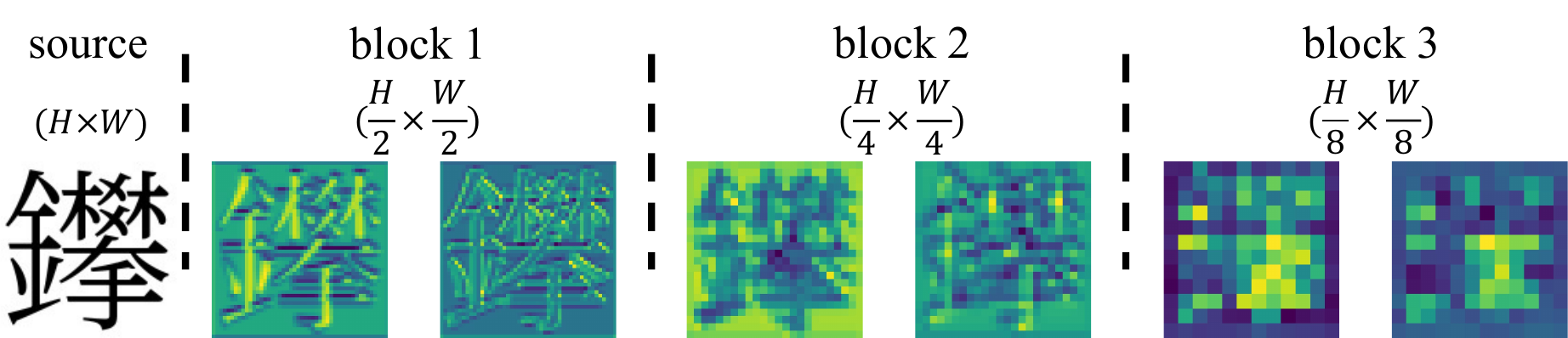}
    \caption{Content features in various blocks.}
    \label{fig:multi-scale_content_feature}
\end{figure}

Generating complex characters has always been a challenging task, and many existing methods only rely on a single-scale content feature, disregarding the intricate details such as strokes and components. As shown in Figure \ref{fig:multi-scale_content_feature}, large-scale features retain lots of detailed information while small-scale features are lack of these.

\begin{figure}[t]
    \centering
    \includegraphics[width=1.0\columnwidth]{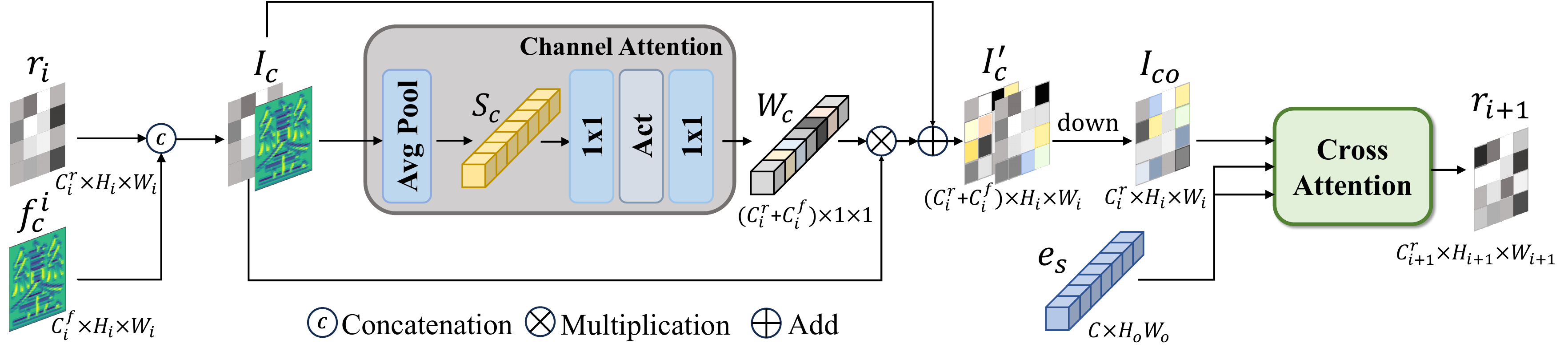}
    \caption{Multi-scale Content Aggregation.}
    \label{fig:MCA}
\end{figure}

Therefore, we employ a Multi-scale Content Aggregation (MCA) block, injecting global and local content features across different scales into the UNet of our diffusion model. Specifically, the source image $\boldsymbol{x}_c$ is first embedded by the content encoder $E_{c}$, obtaining multi-scale content features $\boldsymbol{F}_{c}=\{f_{c}^{1}, f_{c}^{2}, f_{c}^{3}\}$ from different layers. 
Together with the style embedding $e_{s}$ encoded by the style encoder $E_{s}$, each content feature $f_{c}^{i}$ is injected into the UNet through three MCA modules respectively. As illustrated in Figure \ref{fig:MCA}, the content feature $f_{c}^{i}$ is concatenated with the previous UNet block feature $r_{i}$, resulting in a channel-informative feature $I_{c}$. To enhance the capability of adaptive selective channel fusion, we apply a channel attention \cite{hu2018squeeze} on $I_{c}$, in which an average pooling, two $1\times 1$ convolutions and an activation function are employed. The attention results in a global channel-aware vector $W_{c}$, which is used to weight the channel-informative feature $I_{c}$ via channel-wise multiplication. Then, after a residual connection, we employ a $1\times 1$ convolution to reduce the channel number of ${I}'_{c}$, obtaining the output $I_{co}$. Lastly, we apply a cross-attention module to insert the style embedding $e_{s}$, in which $e_{s}$ is employed as Key and Value, while $I_{co}$ is employed as Query.

\subsubsection{Reference-Structure Interaction (RSI)} There exists structural differences (\textit{e.g.}, font size) between the source image and the target image.  To address this issue, we propose a Reference-Structure Interaction (RSI) block that employs deformable convolutional networks (DCN) \cite{dai2017deformable} to conduct structural deformation on the skip connection of UNet. In contrast to \cite{xie2021dg}, our conditional model directly extracts structural information from the reference features to obtain the deformation offset $\delta_{offset}$ for DCN. 

Specifically, the reference image $\boldsymbol{x}_s$ is first passed through the content encoder $E_{c}$ to obtain the structure maps $\boldsymbol{F}_{s}=\{f_{s}^{1}, f_{s}^{2}\}$, and each $f_{s}^{i}$ is as the input to both RSI modules respectively.
There exists misalignment in the spatial position between the UNet feature and the reference feature. Therefore, instead of applying CNN to obtain the offset $\delta_{offset}$ in traditional DCN, we introduce a cross-attention to enable long-distance interactions. The interaction process can be summarized in Equation \ref{equ:RSI}: $r_{i}$ is the UNet feature. And the essential element of this process involves leveraging the UNet feature $r_{i}$ and structure map $f_{s}^{i}$ in a softmax operation, which primarily calculates the region of interest relative to each query position.
\begin{gather}
    S_{s}\in \mathbb{R}^{C_{f}^{i}\times H_{i}W_{i}} = flatten(f_{s}^{i}), \notag \\
    S_{r}\in \mathbb{R}^{C_{r}^{i}\times H_{i}W_{i}} = flatten(r_{i}), \notag \\
    Q = \Phi_{q}(S_{s}),\hspace{5pt} K = \Phi_{k}(S_{r}),\hspace{5pt} V = \Phi_{v}(S_{r}), \notag \\
    F_{attn} = softmax(\frac{QK^{T}}{\sqrt{d_{k}}})V,\hspace{6pt} \delta_{offset} = FFN(F_{attn}), \notag \\
    I_{R} = DCN(r_{i}, \delta_{offset}),
\label{equ:RSI}
\end{gather}
where $\Phi_{q}$, $\Phi_{k}$, $\Phi_{v}$ are linear projections, and $FFN$ denotes the feed forward network. $I_{R}$ is the output of RSI.

\subsection{Style Contrastive Refinement}
\noindent One purpose of font generation is to achieve the intended style imitating effect, regardless of the variations of style between the source and the reference. A novel strategy is to find a suitable style representation and further provide feedback to our model. Therefore, we propose a Style Contrastive Refinement (SCR) module, a font style representation learning module that disentangles style from a group of samples images and incorporates a style contrastive loss to supervise our diffusion model, ensuring the generated style aligns with the target at the global and local level.

The architecture of SCR is shown on the right of Figure \ref{fig:framework}, which consists of a style extractor. Inspired by \cite{zhang2022domain}, a VGG network is employed to embed the font image in the extractor. To capture both global and local style characteristics effectively, we select N layers of feature maps $\boldsymbol{F}_{v}=\{f_{v}^{0}, f_{v}^{1}, ..., f_{v}^{N}\}$ from VGG network, utilizing them as input to a style projector. The projector applies an average pooling and a maximum pooling to extract different global channel features separately, and then concatenates both of them channel-wise, resulting in the features $\boldsymbol{F}_{g}=\{f_{g}^{0}, f_{g}^{1}, ..., f_{g}^{N}\}$. Finally, after several linear projections, style vectors $\boldsymbol{V}=\{v^{0}, v^{1}, ..., v^{N}\}$ are obtained.

The style vectors $\boldsymbol{V}$ can provide supervising signals to the diffusion model and guide it to imitate style. Therefore, we adopt a contrastive learning strategy, in which we leverage a pre-trained SCR and incorporate a style contrastive loss $\mathcal{L}_{sc}$ to supervise whether the style of the generated sample $\boldsymbol{x}_{0}$ is consistent with the target style and distinguishable from negative styles. To ensure content-irrelevance and style-relevance, we choose the target image as the positive sample and select $K$ negative samples that are with different styles but the same content, rather than directly considering the rest of the chosen target sample as negatives. Therefore, the supervision of SCR can be summarized as follows:
\begin{small}
\begin{gather}        
    \boldsymbol{V}_{0} = Extrac(\boldsymbol{x}_{0}), \hspace{5pt} \boldsymbol{V}_{p} = Extrac(\boldsymbol{x}_{p}), \hspace{5pt} \boldsymbol{V}_{n} = Extrac(\boldsymbol{x}_{n}) \notag \\
    \mathcal{L}_{sc}=-\sum_{l=0}^{N-1}log\frac{exp(v_{0}^{l}\cdot v_{p}^{l}/\tau)}{exp(v_{0}^{l}\cdot v_{p}^{l}/\tau)+\sum_{i=1}^{K}exp(v_{0}^{l}\cdot v_{n_{i}}^{l}/\tau)},
\end{gather}
\end{small}
where $Extrac$ represents the style extractor. $K$ is the number of negative samples. $\boldsymbol{V}_{0}$, $\boldsymbol{V}_{p}$ and $\boldsymbol{V}_{n}$ denote the style vectors of generated, positive and negative samples respectively, and $v_{0}^{l}$, $v_{p}^{l}$, $v_{n_{i}}^{l}$ denotes the $l$-th generated, positive and negative layer vector respectively. $\tau$ is a temperature hyper-parameter and set as $0.07$. The pre-training details of SCR are listed in Appendix.

To enhance the robustness of style imitation, we apply an \textit{augmentation strategy} on the positive target sample, which includes random cropping and random resizing.

\subsection{Training Objective}
\noindent Our training adopts a coarse-to-fine two-phase strategy.
\subsubsection{Phase 1} During phase 1, we optimize FontDiffuser mainly with the standard MSE diffusion loss, excluding the SCR module. This ensures that our generator acquires the fundamental capability for font reconstruction:
\begin{align}
    \mathcal{L}_{total}^{1} = \mathcal{L}_{MSE}+\lambda_{cp}^{1} \mathcal{L}_{cp}+\lambda_{off}^{1} \mathcal{L}_{offset},
\end{align}
in which,
\begin{gather}
    \mathcal{L}_{MSE} = \left \| \boldsymbol{\epsilon} - \boldsymbol{\epsilon}_{\theta} (\boldsymbol{x}_{t}, t, \boldsymbol{x}_{c}, \boldsymbol{x}_{s}) \right \|^{2}, \\
    \mathcal{L}_{cp} = \sum_{l=1}^{L} \left \|\mathcal{VGG}_{l}(\boldsymbol{x}_{0})-\mathcal{VGG}_{l}(\boldsymbol{x}_{target}) \right \|, \\
    \mathcal{L}_{offset}=mean(\left \| \delta_{offset} \right \|),
\end{gather}
where $\mathcal{L}_{total}^{1}$ represents the total loss in phase 1. $\mathcal{VGG}_{l}(\cdot)$ is the layer feature encoded by VGG and $L$ is the number of the chosen layers. $\mathcal{L}_{cp}$ is used to penalize the content misalignment between generated VGG features of $\boldsymbol{x}_{0}$ and the corresponding $\boldsymbol{x}_{target}$ target features. The offset loss $\mathcal{L}_{offset}$ is used to constrain the offset in our RSI module and $mean$ is the averaging process. $\lambda_{cp}^{1}=0.01$ and $\lambda_{off}^{1}=0.5$.

\subsubsection{Phase 2}In phase 2, we implement the SCR module, incorporating the style contrastive loss, to provide style imitation guidance to the diffusion model at the global and local levels. Thus our conditional diffusion model in phase 2 is optimized by:
\begin{align}
    \mathcal{L}_{total}^{2} = \mathcal{L}_{MSE}+\lambda_{cp}^{2} \mathcal{L}_{cp}+\lambda_{off}^{2} \mathcal{L}_{offset}+\lambda_{sc}^{2} \mathcal{L}_{sc},
\end{align}
where $\mathcal{L}_{total}^{2}$ represents the total loss in phase 2. The hyper-parameters $\lambda_{cp}^{2}=0.01$, $\lambda_{off}^{2}=0.5$ and $\lambda_{sc}^{2}=0.01$.

\section{Experiment}
\subsection{Datasets and Evaluation Metrics}
We collect a Chinese font dataset of 424 fonts. We randomly select 400 fonts (referred to as ``seen fonts") with 800 Chinese characters (referred to as ``seen characters") as training set. We evaluate methods on two test sets: one includes 100 randomly selected seen fonts, which contains 272 characters that were not seen during training (referred to as ``SFUC"), and the other test set consists of 24 unseen fonts and 300 unseen characters (referred to as ``UFUC"). The categorization details of three levels of complexity are in Appendix. Moreover, we additionally conduct a comparison on 24 unseen fonts and 800 seen characters (referred to as ``UFSC").

For quantitative evaluation, we adopt FID, SSIM, LPIPS, and L1 loss metrics. Pixel-level metrics SSIM and L1 loss are employed to measure the per-pixel consistency between generated samples and target samples. Moreover, LPIPS \cite{zhang2018unreasonable} and FID \cite{heusel2017gans} are perceptual metrics, which are closer to human visual perception. Furthermore, we conduct a \textit{user study} to assess the subjective quality of images. We randomly select 30 seen fonts from SFUC and 20 unseen fonts from UFUC. In each font, we randomly select 6 characters (2 characters per complexity). In total, 25 participants are asked to choose the best from the results of all methods.

\subsection{Implementation Details}
We train FontDiffuser using AdamW optimizer with $\beta_{1} = 0.9$ and $\beta_{2} = 0.999$. The image size is set as $96$. Moreover, following \cite{ho2022classifier}, we simply drop out the source image and the reference image with the probability of $0.1$.
In phase 1, we train the model with a batch size of $16$ and a total step of 440000. The learning rate is set as $1e-4$ with linear schedule. 
In phase 2, the learning rate is set as $1e-5$ and is fixed as constant. We train with a batch size of $16$, a total step of 30000, and negative samples of $16$.
The experiments are conducted on a single RTX 3090 GPU.

During sampling, we adopt a classifier-free guidance strategy \cite{ho2022classifier} to amplify the effect of the conditions $\boldsymbol{x}_{c}$ and $\boldsymbol{x}_{s}$. We set the unconditional content image and unconditional style image to pixel 255 as $\boldsymbol{\emptyset}$, and our sampling strategy can be formulated as:
\begin{small}
\begin{align}
    \boldsymbol{\epsilon}_{\theta} (\boldsymbol{x}_{t}, t, \boldsymbol{x}_{c}, \boldsymbol{x}_{s})=(1-s)\boldsymbol{\epsilon}_{\theta} (\boldsymbol{x}_{t}, t, \boldsymbol{\emptyset}, \boldsymbol{\emptyset})+s\boldsymbol{\epsilon}_{\theta}(\boldsymbol{x}_{t}, t, \boldsymbol{x}_{c}, \boldsymbol{x}_{s}),
\end{align}
\end{small}
where $s$ is the guidance scale and is set as $7.5$ in the experiments. To speed up sampling, we use the DPM-Solver++ sampler \cite{lu2022dpm} with only 20 inference steps. 

\begin{table*}[t]
\centering
\resizebox{16.3cm}{!}{%
\begin{tabular}{@{}lccccccccccccccccccl@{}}
\toprule
\multirow{2}{*}{}                          & \multirow{2}{*}{Model}   & \multirow{2}{*}{Venue}    & \multicolumn{4}{c}{Easy}                                                                    & \multicolumn{4}{c}{Medium}                                                                  & \multicolumn{4}{c}{Hard}                                                                    & \multicolumn{4}{c}{Average}                                                                & \multicolumn{1}{c}{\multirow{2}{*}{\begin{tabular}[c]{@{}c@{}}User\\ ($\%$)\end{tabular}}} \\ \cmidrule(lr){4-19}
                                           &              &                & FID$\downarrow$              & SSIM$\uparrow$            & LPIPS$\downarrow$           & \multicolumn{1}{c}{L1$\downarrow$}           & FID$\downarrow$              & SSIM$\uparrow$            & LPIPS$\downarrow$           & \multicolumn{1}{c}{L1$\downarrow$}           & FID$\downarrow$              & SSIM$\uparrow$            & LPIPS$\downarrow$           & \multicolumn{1}{c}{L1$\downarrow$}           & FID$\downarrow$             & SSIM$\uparrow$            & LPIPS$\downarrow$           & \multicolumn{1}{c}{L1$\downarrow$}           & \multicolumn{1}{c}{}                                                                      \\ \midrule
\multicolumn{1}{l|}{\multirow{8}{*}{\begin{turn}{90}SFUC\end{turn}}} & \multicolumn{1}{l|}{FUNIT}     & \multicolumn{1}{c|}{ICCV2019}   & 11.3390          & 0.4342          & 0.1985          & \multicolumn{1}{c|}{0.3888}          & 11.0158          & 0.3516          & 0.2144          & \multicolumn{1}{c|}{0.4474}          & 17.9055          & 0.3271          & 0.2374          & \multicolumn{1}{c|}{0.4648}          & 9.6683          & 0.3755          & 0.2146          & \multicolumn{1}{c|}{0.4305}          & 1.30                                                                                      \\
\multicolumn{1}{l|}{}                      & \multicolumn{1}{l|}{LF-Font}   & \multicolumn{1}{c|}{AAAI2021}  & 18.0056          & 0.4914          & 0.1770          & \multicolumn{1}{c|}{0.3325}          & 25.7196          & 0.3833          & 0.2048          & \multicolumn{1}{c|}{0.4184}          & 39.6788          & 0.3444          & 0.2343          & \multicolumn{1}{c|}{0.4511}          & 22.7387         & 0.4127          & 0.2024          & \multicolumn{1}{c|}{0.3955}          & 0.37                                                                                      \\
\multicolumn{1}{l|}{}                      & \multicolumn{1}{l|}{DG-Font}  & \multicolumn{1}{c|}{CVPR2021}  & 20.4848          & 0.4613          & 0.2111          & \multicolumn{1}{c|}{0.3610}          & 24.4368          & 0.3831          & 0.2354          & \multicolumn{1}{c|}{0.4146}          & 29.5987          & 0.3444          & 0.2614          & \multicolumn{1}{c|}{0.4430}          & 21.1623         & 0.4016          & 0.2333          & \multicolumn{1}{c|}{0.4024}          & 5.28                                                                                      \\
\multicolumn{1}{l|}{}                      & \multicolumn{1}{l|}{MX-Font}  & \multicolumn{1}{c|}{ICCV2021}  & 12.4251          & 0.4693          & \underline{0.1688}          & \multicolumn{1}{c|}{0.3511}          & 11.2868          & 0.3790          & \underline{0.1784}          & \multicolumn{1}{c|}{0.4184}          & \underline{14.1061}          & 0.3338          & \underline{0.1964}          & \multicolumn{1}{c|}{0.4546}          & 10.2200         & 0.4002          & \underline{0.1796}          & \multicolumn{1}{c|}{0.4033}          & 11.39                                                                                     \\
\multicolumn{1}{l|}{}                      & \multicolumn{1}{l|}{Fs-Font} & \multicolumn{1}{c|}{CVPR2022}  & 27.1983          & 0.4282          & 0.2258          & \multicolumn{1}{c|}{0.3869}          & 27.9421          & 0.3425          & 0.2394          & \multicolumn{1}{c|}{0.4536}          & 36.9010          & 0.3091          & 0.2621          & \multicolumn{1}{c|}{0.4800}          & 25.9870         & 0.3651          & 0.2404          & \multicolumn{1}{c|}{0.4361}          & 0.83                                                                                      \\
\multicolumn{1}{l|}{}                      & \multicolumn{1}{l|}{CG-GAN}  & \multicolumn{1}{c|}{CVPR2022}  & \textbf{8.2271}  & 0.4692          & 0.1816          & \multicolumn{1}{c|}{0.3582}          & \textbf{9.1112}           & 0.3755          & 0.1952          & \multicolumn{1}{c|}{0.4280}          & 14.1878          & 0.3396          & 0.2173          & \multicolumn{1}{c|}{0.4570}          & \underline{7.7862}          & 0.4004          & 0.1961          & \multicolumn{1}{c|}{0.4100}          & \underline{32.87}                                                                                     \\
\multicolumn{1}{l|}{}                      & \multicolumn{1}{l|}{CF-Font} & \multicolumn{1}{c|}{CVPR2023}  & 14.0800          & \underline{0.4924}          & 0.2015          & \multicolumn{1}{c|}{\underline{0.3224}}          & 13.9623          & \underline{0.4006}          & 0.2347          & \multicolumn{1}{c|}{\underline{0.3929}}          & 16.8435          & \underline{0.3662}          & 0.2634          & \multicolumn{1}{c|}{\underline{0.4197}}          & 12.1268         & \underline{0.4253}          & 0.2301          & \multicolumn{1}{c|}{\underline{0.3741}}          & 1.11                                                                                      \\
\multicolumn{1}{l|}{}                      & \multicolumn{1}{l|}{Ours}  & \multicolumn{1}{c|}{-}  & \underline{8.5089}           & \textbf{0.5370} & \textbf{0.1316} & \multicolumn{1}{c|}{\textbf{0.2901}} & \underline{9.4580}           & \textbf{0.4462} & \textbf{0.1411} & \multicolumn{1}{c|}{\textbf{0.3623}} & \textbf{11.1475} & \textbf{0.4033} & \textbf{0.1562} & \multicolumn{1}{c|}{\textbf{0.3986}} & \textbf{7.6985} & \textbf{0.4682} & \textbf{0.1416} & \multicolumn{1}{c|}{\textbf{0.3454}} & \textbf{46.85}                                                                            \\ \midrule
\multicolumn{1}{l|}{\multirow{8}{*}{\begin{turn}{90}UFUC\end{turn}}} & \multicolumn{1}{l|}{FUNIT}  & \multicolumn{1}{c|}{ICCV2019}  & 14.5517          & 0.4507          & 0.1839          & \multicolumn{1}{c|}{0.3720}          & 16.0900          & 0.3495          & 0.2045          & \multicolumn{1}{c|}{0.4484}          & 25.9712          & 0.2963          & 0.2403          & \multicolumn{1}{c|}{0.4918}          & 13.1426         & 0.3655          & 0.2095          & \multicolumn{1}{c|}{0.4374}          & 2.03                                                                                      \\
\multicolumn{1}{l|}{}                      & \multicolumn{1}{l|}{LF-Font}  & \multicolumn{1}{c|}{AAAI2021}  & 23.9173          & \underline{0.4949}          & 0.1687          & \multicolumn{1}{c|}{\underline{0.3301}}          & 38.6071          & 0.3746          & 0.1997          & \multicolumn{1}{c|}{0.4257}          & 55.4416          & 0.3071          & 0.2370          & \multicolumn{1}{c|}{0.4833}          & 32.8862         & 0.3922          & 0.2018          & \multicolumn{1}{c|}{0.4130}          & \multicolumn{1}{c}{0.10}                                                                  \\
\multicolumn{1}{l|}{}                      & \multicolumn{1}{l|}{DG-Font}  & \multicolumn{1}{c|}{CVPR2021}  & 25.6115          & 0.4788          & 0.1957          & \multicolumn{1}{c|}{0.3450}          & 27.0834          & 0.3803          & 0.2172          & \multicolumn{1}{c|}{0.4165}          & 32.7255          & 0.3254          & 0.2421          & \multicolumn{1}{c|}{0.4561}          & 22.7077         & 0.3948          & 0.2183          & \multicolumn{1}{c|}{0.4059}          & \multicolumn{1}{c}{8.99}                                                                  \\
\multicolumn{1}{l|}{}                      & \multicolumn{1}{l|}{MX-Font}  & \multicolumn{1}{c|}{ICCV2021}  & 14.9232          & 0.4808          & \underline{0.1552}          & \multicolumn{1}{c|}{0.3408}          & \underline{14.0944}          & 0.3786          & \underline{0.1625}          & \multicolumn{1}{c|}{0.4195}          & \underline{16.3962}          & 0.3189          & \underline{0.1783}          & \multicolumn{1}{c|}{0.4689}          & \underline{10.7689}         & 0.3928          & \underline{0.1653}          & \multicolumn{1}{c|}{0.4098}          & 14.11                                                                                     \\
\multicolumn{1}{l|}{}                      & \multicolumn{1}{l|}{Fs-Font}  & \multicolumn{1}{c|}{CVPR2022}  & 42.7799          & 0.4524          & 0.2100          & \multicolumn{1}{c|}{0.3646}          & 43.6933          & 0.3495          & 0.2282          & \multicolumn{1}{c|}{0.4448}          & 49.3266          & 0.2973          & 0.2565          & \multicolumn{1}{c|}{0.4869}          & 38.7702         & 0.3664          & 0.2315          & \multicolumn{1}{c|}{0.4321}          & \multicolumn{1}{c}{1.26}                                                                  \\
\multicolumn{1}{l|}{}                      & \multicolumn{1}{l|}{CG-GAN}  & \multicolumn{1}{c|}{CVPR2022}   & \underline{14.1445}          & 0.4887          & 0.1677          & \multicolumn{1}{c|}{0.3369}          & 14.4114          & 0.3793          & 0.1831          & \multicolumn{1}{c|}{0.4173}          & 26.8940          & 0.3114          & 0.2120          & \multicolumn{1}{c|}{0.4710}          & 12.9301         & 0.3931          & 0.1876          & \multicolumn{1}{c|}{0.4084}          & \multicolumn{1}{c}{\underline{20.97}}                                                                 \\
\multicolumn{1}{l|}{}                      & \multicolumn{1}{l|}{CF-Font}  & \multicolumn{1}{c|}{CVPR2023}  & 22.0913          & 0.4841          & 0.1901          & \multicolumn{1}{c|}{0.3322}          & 24.5819          & \underline{0.3897}          & 0.2180          & \multicolumn{1}{c|}{\underline{0.4046}}          & 25.8287          & \textbf{0.3434} & 0.2461          & \multicolumn{1}{c|}{\textbf{0.4420}} & 19.6929         & \underline{0.4057}          & 0.2180          & \multicolumn{1}{c|}{\underline{0.3929}}          & \multicolumn{1}{c}{5.41}                                                                  \\
\multicolumn{1}{l|}{}                      & \multicolumn{1}{l|}{Ours}  & \multicolumn{1}{c|}{-}   & \textbf{12.8973} & \textbf{0.5080} & \textbf{0.1418} & \multicolumn{1}{c|}{\textbf{0.3175}} & \textbf{11.6271} & \textbf{0.4117} & \textbf{0.1468} & \multicolumn{1}{c|}{\textbf{0.3926}} & \textbf{13.1228} & \underline{0.3420}          & \textbf{0.1600} & \multicolumn{1}{c|}{\underline{0.4508}}          & \textbf{8.5352} & \textbf{0.4206} & \textbf{0.1496} & \multicolumn{1}{c|}{\textbf{0.3870}} & \multicolumn{1}{c}{\textbf{47.15}}                                                        \\ \bottomrule
\end{tabular}%
}
\caption{Quantitative Results on SFUC and UFUC. `User' denotes the user study. `Average' and the user study is evaluated on all characters of three levels of complexity. The bold indicates the state-of-the-art and the underline indicates the second best.}
\label{tab:quantitative_sfuc_ufuc}
\end{table*}

\subsection{Comparison with State-of-the-Art Method}
\noindent We compare our method with seven state-of-the-art methods: one image-to-image translation method (FUNIT \cite{liu2019few}) and six Chinese font generation methods (LF-Font \cite{park2021few}, MX-Font \cite{park2021multiple}, DG-Font \cite{xie2021dg}, CG-GAN \cite{kong2022look}, Fs-Font \cite{tang2022few}, and CF-Font \cite{wang2023cf}). Additionally, we compare with Diff-Font \cite{he2022diff} on Unseen Font Seen Character (UFSC). For a fair comparison, we use the font of Song as the source, and all methods are trained based on their official codes. 

\subsubsection{Quantitative comparison} The quantitative results are presented in Table \ref{tab:quantitative_sfuc_ufuc}. FontDiffuser achieves the best performance across all matrices at average level, showing a significant gap compared to other methods on both SFUC and UFUC. It indicates that FontDiffuser can generate fonts that are visually closer to human perception. At easy and medium levels, though FID in SFUC ranks second, FontDiffuser outperforms other methods in the remaining metrics, particularly the perceptual matrix LPIPS. At hard level, our method performs the best in SFUC and achieves the best FID and LPIPS scores in UFUC. It should be noted that SSIM and L1 loss are pixel-level metrics, which may not directly reflect the overall performance. For instance, an impressive visual result may not perfectly match the target pixel to pixel. The hard-level results demonstrate the advantage of FontDiffuser in generating complex characters. Furthermore, as shown in Table \ref{tab:quantitative_ufsc}, FontDiffuser achieves state-of-the-art performance on UFSC. Notably, Diff-Font \cite{he2022diff} is only capable of generating seen characters, and our method also outperforms it by a significant margin.

\begin{table}[]
    \centering
    \resizebox{6cm}{!}{%
    \begin{tabular}{@{}l|c|cccc@{}}
    \toprule
    Model   &Venue  & FID$\downarrow$             & SSIM$\uparrow$            & LPIPS$\downarrow$           & L1$\downarrow$              \\ \midrule
    LF-Font &ICCV2019  & 18.6368         & \underline{0.4823}          & 0.1688          & \underline{0.3400}          \\
    DG-Font &CVPR2021  & 19.8079         & 0.4532          & 0.2047          & 0.3646          \\
    MX-Font &ICCV2021  & 9.3238          & 0.4605          & \underline{0.1603}          & 0.3571          \\
    Fs-Font &CVPR2022  & 31.3986         & 0.4270          & 0.2160          & 0.3855          \\
    CG-GAN  &CVPR2022  & \underline{7.7232}          & 0.4655          & 0.1721          & 0.3544          \\
    CF-Font &CVPR2023  & 14.2027         & 0.4396          & 0.2139          & 0.3713          \\
    Diff-Font &- & 12.0809         & 0.4192          & 0.2022          & 0.3877          \\
    Ours  &-   & \textbf{7.6708}  & \textbf{0.4942} & \textbf{0.1426} & \textbf{0.3279} \\ \bottomrule
    \end{tabular}%
    }
    \caption{Quantitative Results on UFSC.}
    \label{tab:quantitative_ufsc}
    \end{table}

\begin{figure}[]
    \centering
    \includegraphics[width=0.6\columnwidth]{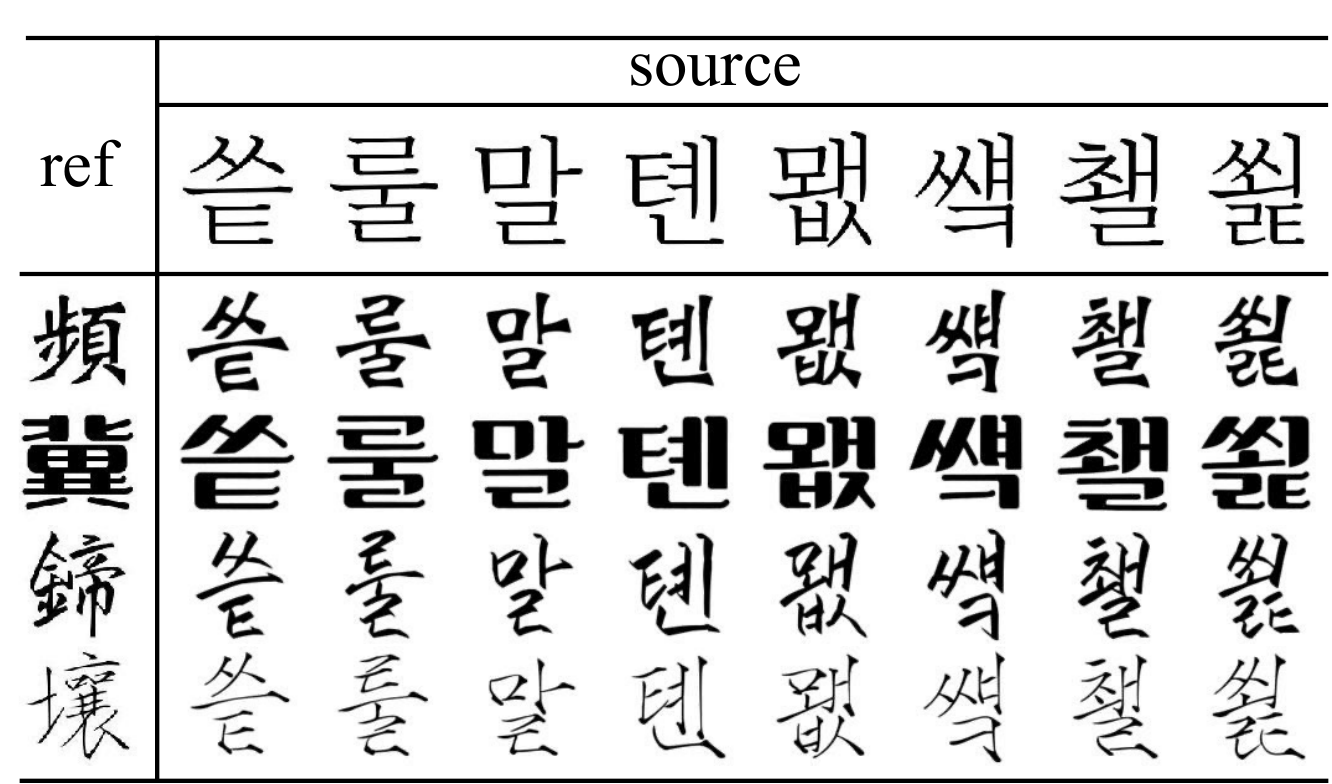}
    \caption{Cross-lingual generation (Chinese to Korean).}
    \label{fig:korean_gen}
\end{figure}

\subsubsection{Qualitative comparison} 
\begin{figure*}[]
    \centering
    \includegraphics[width=0.75\textwidth]{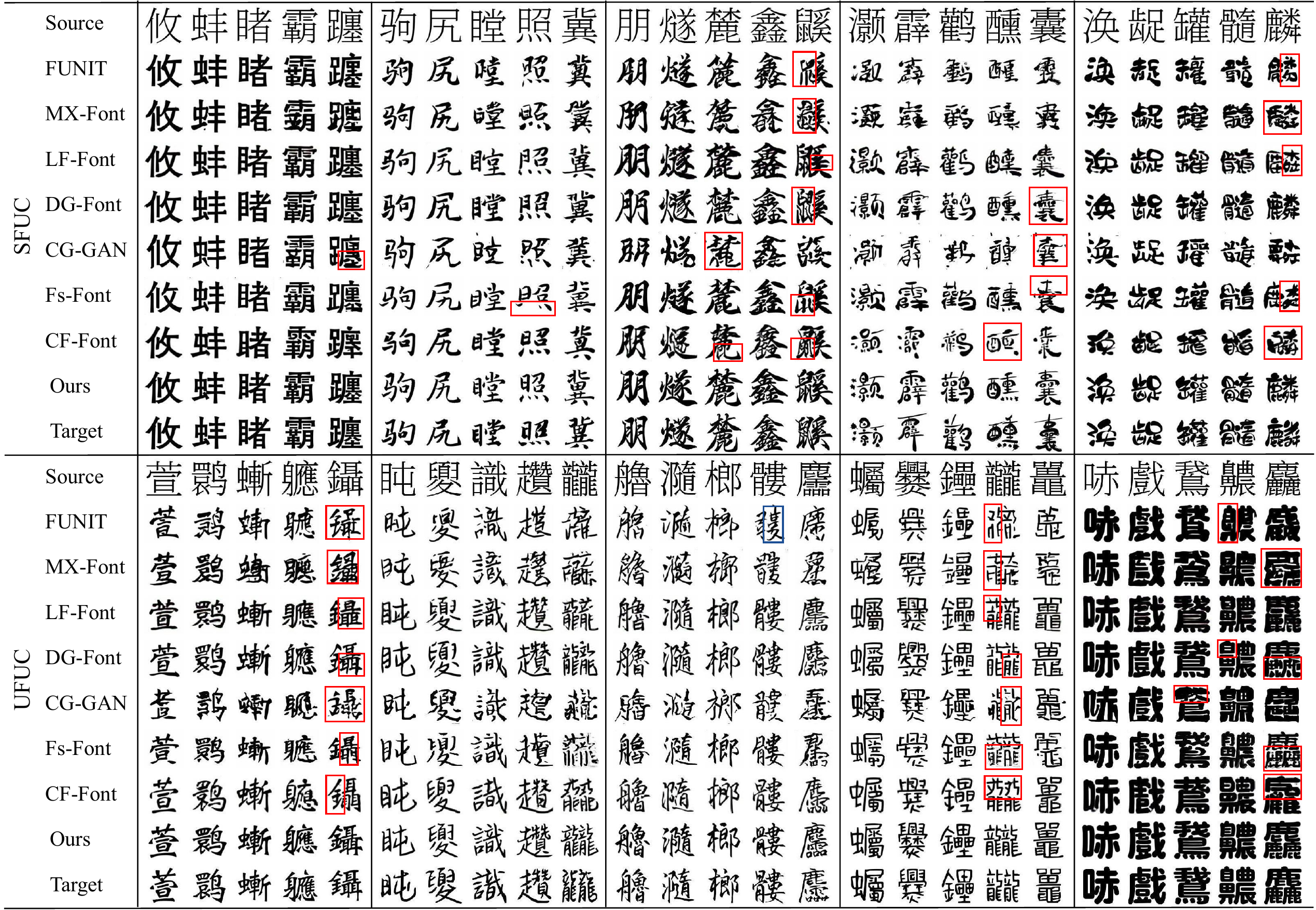}
    \caption{Qualitative comparison on SFUC and UFUC. Red boxes highlight the failures of other methods.}
    \label{fig:result_vis}
\end{figure*}

In Figure \ref{fig:result_vis}, we provide visualizations of the results on SFUC and UFUC, which intuitively reflect the visual effects of different methods. FontDiffuser consistently generates high-quality results and performs better in terms of content preservation, style consistency, and structural correctness compared with other state-of-the-art methods. Particularly, our method demonstrates significant superiority in generating complex characters and handling large variations in style transfer, while other methods still exhibit issues such as missing strokes, artifacts, blurriness, layout errors, and style inconsistency. We also present some cross-lingual generation samples (Chinese to Korean) in Figure \ref{fig:korean_gen}, which are generated by our method. It demonstrates that FontDiffuser is flexible in generating for other languages and exhibits cross-domain capability though our model is trained by Chinese dataset.

\begin{table}[]
    \centering
    \resizebox{5cm}{!}{%
    \begin{tabular}{@{}ccccccc@{}}
    \toprule
    \multicolumn{3}{c}{Module}     & \multirow{2}{*}{FID$\downarrow$} & \multirow{2}{*}{SSIM$\uparrow$} & \multirow{2}{*}{LPIPS$\downarrow$} & \multirow{2}{*}{L1$\downarrow$} \\
    M & R & S                      &                      &                       &                        &                         \\ \midrule
    \usym{2717} & \usym{2717} & \multicolumn{1}{c|}{\usym{2717}} & \underline{8.1153}               & 0.4112                & 0.1526                 & 0.3955                  \\
    \usym{2713} & \usym{2717} & \multicolumn{1}{c|}{\usym{2717}} & \textbf{7.8419}      & 0.4114                & 0.1511                 & 0.3954                  \\
    \usym{2713} & \usym{2713} & \multicolumn{1}{c|}{\usym{2717}} & 8.4427               & \underline{0.4137}                & \underline{0.1506}                 & \underline{0.3925}                  \\
    \usym{2713} & \usym{2713} & \multicolumn{1}{c|}{\usym{2713}} & 8.5352               & \textbf{0.4206}       & \textbf{0.1496}        & \textbf{0.3870}         \\ \bottomrule
    \end{tabular}%
    }
    \caption{Effectiveness of different modules. M, R, and S represent MCA, RSI, and SCR respectively. The first row represents the baseline.}
\label{tab:effectiveness_modules}
\end{table}

\subsection{Ablation Studies}
In this section, we conduct several ablation studies to analyze the performance of our proposed modules and strategies. The experiments are tested on the unseen font unseen characters (UFUC) at average level.
\subsubsection{Effectiveness of different modules} 
\begin{figure}[]
    \centering
    \includegraphics[width=0.7\columnwidth]{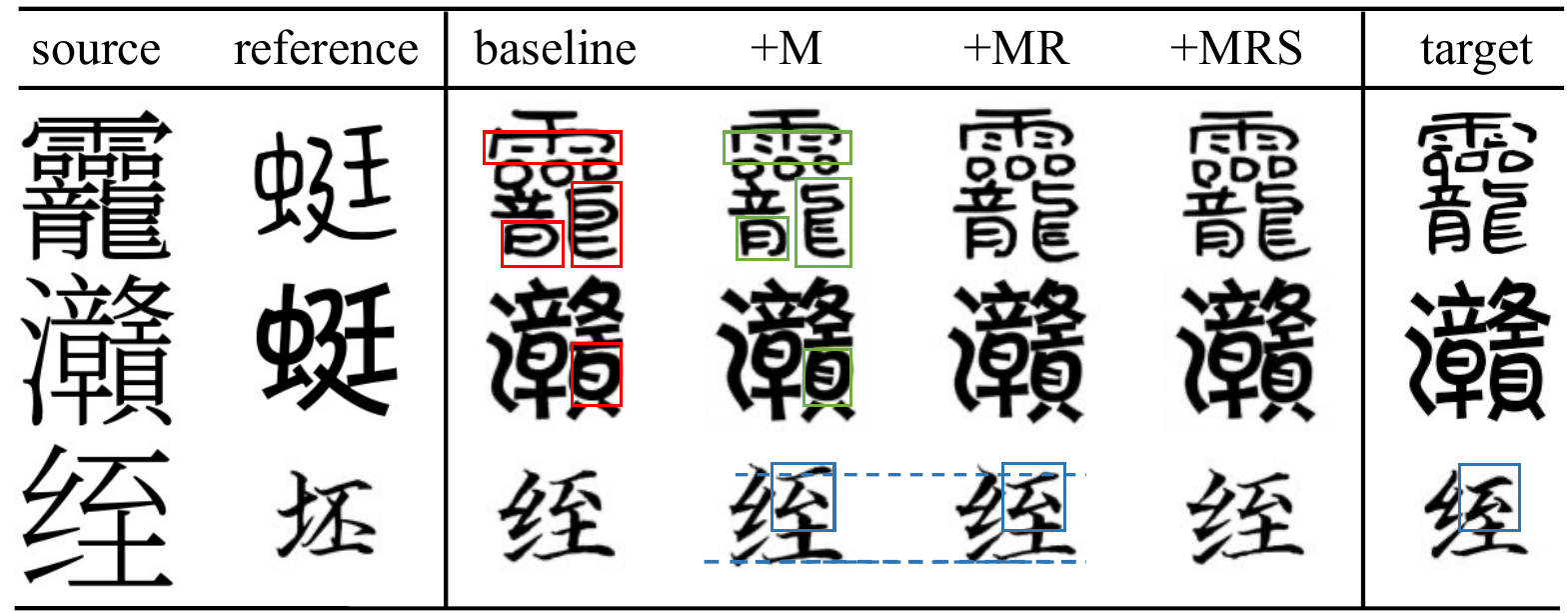}
    \caption{Visualization of different modules. M, R, and S represent MCA, RSI, and SCR respectively. Red boxes represent the missing strokes while green represents the corresponding improvements. Blue denotes structural promotion.}
    \label{fig:ablation_vis}
\end{figure}

We separate the proposed MCA, RSI, and SCR, and progressively add them to the baseline. The baseline concatenates the content image with $\boldsymbol{x}_{t}$ as the input of UNet. Table \ref{tab:effectiveness_modules} shows that the quantitative results of these three modules are improved in terms of SSIM, LPIPS, and L1 loss, except for FID. Additionally, these modules also contribute to visual enhancements, as shown in Figure \ref{fig:ablation_vis}. For example, in the first row of Figure \ref{fig:ablation_vis}, the issue of missing strokes in the baseline is mitigated by the incorporation of the MCA module.

\subsubsection{Effectiveness of augmentation strategy in SCR} We investigate the advantage of the proposed augmentation strategy in SCR, in which FontDiffuser is trained with and without augmentation strategy during the training phase 2. As shown in Table \ref{tab:effectiveness_aug}, it clearly demonstrates that the augmentation strategy boosts the generation performance in terms of SSIM, LPIPS, and L1 loss.

\begin{table}[]
\centering
\resizebox{6.3cm}{!}{%
\begin{tabular}{@{}c|cccl@{}}
\toprule
Method  & FID$\downarrow$             & SSIM$\uparrow$            & LPIPS$\downarrow$           & \multicolumn{1}{c}{L1$\downarrow$} \\ \midrule
w/o augmentation & \textbf{8.1758} & 0.4172          & 0.1504          & 0.3900                     \\
augmentation     & 8.5352          & \textbf{0.4206} & \textbf{0.1496} & \textbf{0.3870}            \\ \bottomrule
\end{tabular}%
}
\caption{Effectiveness of augmentation strategy in SCR.}
\label{tab:effectiveness_aug}
\end{table}

\subsubsection{Comparison between cross-attention interaction and CNN in RSI} We conduct a comparative analysis between cross-attention interaction and CNN interaction in RSI. The results in Table \ref{tab:comparison_attn} show that the cross-attention interaction in RSI outperforms the CNN-based in all matrices, showcasing the superiority of our proposed method.

\begin{table}[]
\centering
\resizebox{6cm}{!}{%
\begin{tabular}{@{}c|cccl@{}}
\toprule
Method     & FID$\downarrow$             & SSIM$\uparrow$            & LPIPS$\downarrow$           & \multicolumn{1}{c}{L1$\downarrow$} \\ \midrule
CNN        & 9.1659          & 0.4130          & 0.1537          & 0.3932                     \\
cross-attention & \textbf{8.5352} & \textbf{0.4206} & \textbf{0.1496} & \textbf{0.3870}            \\ \bottomrule
\end{tabular}%
}
\caption{Comparison between cross-attention and CNN.}
\label{tab:comparison_attn}
\end{table}

\subsubsection{Others} Additionally, we further discuss more ablation studies in Appendix, including the influence of negative samples for style contrastive loss, the influence of VGG layer features in SCR, and the influence of guidance scales.

\begin{figure}[]
    \centering
    \includegraphics[width=0.75\columnwidth]{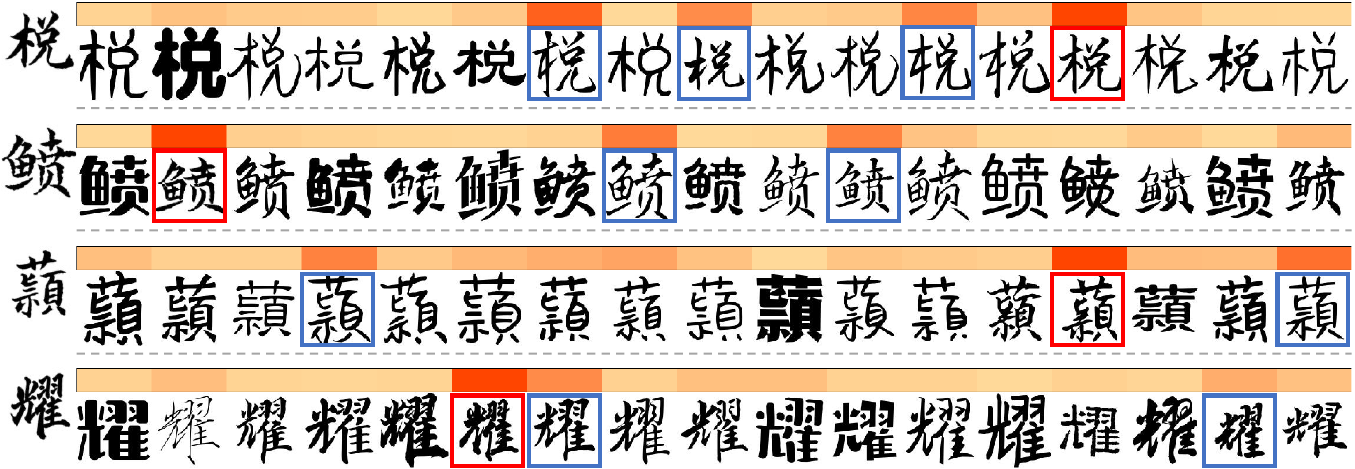}
    \caption{Visualization of SCR contrastive score. The left column represents the generated samples. Each row corresponds to the chosen samples. Red boxes highlight the target while blues highlight samples similar to the generated style. And the darker color in color bars indicates a larger contrastive score while the lighter indicates a smaller one.}
    \label{fig:scr_sim}
\end{figure}

\subsection{Visualization of SCR contrastive score}
We provide visualization of the SCR contrastive score in Figure \ref{fig:scr_sim}, which demonstrates that SCR can effectively distinguish the target from a group of samples, even though some of them exhibit similar styles. By combining SCR with style contrastive loss, we observe that SCR can refine the generated style through a learning-by-contrast manner.

\section{Conclusion}
In this paper, we propose a diffusion-based image-to-image font generation method, called FontDiffuser, which excels in generating complex characters and handling large variations in style transfer. Specifically, we propose the MCA block to inject multi-scale content features into our diffusion model, enhancing the preservation of complex characters. Moreover, we propose a novel style representation learning strategy, which implements the SCR module and uses a style contrastive loss to supervise our diffusion model. Additionally, an RSI block is employed to facilitate structural deformation using reference features. Extensive experiments demonstrate that FontDiffuser outperforms the state-of-the-art method on characters of three levels of complexity. Furthermore, FontDiffuser demonstrates its applicability to the cross-lingual font generation task (\textit{e.g.}, Chinese to Korean), highlighting its promising cross-domain capability.

\section{Acknowledgements}
This research is supported in part by National Key Research and Development Program of China  (2022YFC3301703) and Alibaba Innovative Research Foundation (no. 20210975). We thank the support from the Alibaba-South China University of Technology Joint Graduate Education Program.

\bibliography{aaai24}

\clearpage

\section{Method Details}
\subsection{Conditional Diffusion for Font Generation}
In this section, we present more details of our conditional diffusion model, which is conditioned on a source image $\boldsymbol{x}_{c}$ and a single reference image $\boldsymbol{x}_{s}$, and predicts the added noise $\boldsymbol{\epsilon_{\theta}}$. Our diffusion model consists of a content encoder $E_{c}$, a style encoder $E_{s}$, and a UNet.
\subsubsection{Content Encoder $E_{c}$ and Style Encoder $E_{s}$} In our diffusion model, we adopt the content encoder and style encoder from CG-GAN \cite{kong2022look}. Specifically, we only accept the first three blocks as ours in the content encoder.

\subsubsection{UNet} As shown in Table \ref{tab:unet_architecture}, the UNet in FontDiffuser is made up of Conv blocks, Down blocks, Up blocks, Multi-scale Content Aggregation (MCA) blocks, and Style Insertion (SI) blocks. Style Insertion (SI) block employs a cross-attention module to insert the style embedding $e_{s}$ into the UNet. Down block and Up block represent the downsample and upsample blocks respectively. Conv block is the convolution block. The input of the UNet is $\boldsymbol{x}_{t}\in \mathbb{R}^{3\times H\times W}$ and the output is $\boldsymbol{\epsilon}_{t}\in \mathbb{R}^{3\times H\times W}$.

\begin{table}[h]
\centering
\resizebox{\columnwidth}{!}{%
\begin{tabular}{@{}c|ccc@{}}
\toprule
Block                         & \begin{tabular}[c]{@{}c@{}}Block\\ Number\end{tabular} & \begin{tabular}[c]{@{}c@{}}Input\\ Shape\end{tabular} & \begin{tabular}[c]{@{}c@{}}Output\\ Shape\end{tabular} \\ \midrule
Conv block                  & 1                                                      & $3\times H\times W$                                                     & $64\times H\times W$                                                      \\ \midrule
Down block                  & 2                                                      & $64\times H\times W$                                                     & $64\times \frac{H}{2} \times \frac{W}{2}$                                                      \\ \midrule
MCA block & 2                                                      & $64\times \frac{H}{2} \times \frac{W}{2}$                                                     & $128\times \frac{H}{4} \times \frac{W}{4}$                                                      \\ \midrule
MCA block & 2                                                      & $128\times \frac{H}{4} \times \frac{W}{4}$                                                     & $256\times \frac{H}{8} \times \frac{W}{8}$                                                      \\ \midrule
Down block                     & 2                                                      & $256\times \frac{H}{8} \times \frac{W}{8}$                                                     & $512\times \frac{H}{8} \times \frac{W}{8}$                                                      \\ \midrule
MCA block & 1                                                      & $512\times \frac{H}{8} \times \frac{W}{8}$                                                     & $512\times \frac{H}{8} \times \frac{W}{8}$                                                      \\ \midrule
Up block                       & 3                                                      & $512\times \frac{H}{8} \times \frac{W}{8}$                                                     & $256\times \frac{H}{4} \times \frac{W}{4}$                                                      \\ \midrule
SI block         & 3                                                      & $256\times \frac{H}{4} \times \frac{W}{4}$                                                     & $256\times \frac{H}{2} \times \frac{W}{2}$                                                      \\ \midrule
SI block        & 3                                                      & $256\times \frac{H}{2} \times \frac{W}{2}$                                                     & $128\times H \times W$                                                      \\ \midrule
Up block                       & 3                                                      & $128\times H \times W$                                                     & $64\times H \times W$                                                      \\ \midrule
Conv block                   & 1                                                      & $64\times H \times W$                                                     & $3\times H \times W$                                                      \\ \bottomrule
\end{tabular}%
}
\caption{UNet architecture. Style Insertion (SI) block employs a cross-attention module to insert the style embedding $e_{s}$ into the UNet. Down block and Up block represent the downsample and upsample blocks respectively. Conv block is the convolution block. }
\label{tab:unet_architecture}
\end{table}

\subsection{Style Contrastive Refinement}
\subsubsection{Calculation of $\boldsymbol{x}_{0}$ for SCR} Style Contrastive Refinement (SCR) module is employed to supervise our diffusion model whether the style of the generated sample $\boldsymbol{x}_{0}$ is consistent with the target style. Specifically, we calculate the original sample $\boldsymbol{x}_{0}$ at time step $t$ after the model predicts the noise $\boldsymbol{\epsilon}_{\theta}(\boldsymbol{x}_{t}, t, \boldsymbol{x}_{c}, \boldsymbol{x}_{s})$ as:
\begin{align}
    \boldsymbol{x}_{0} = \frac{1}{\sqrt{\bar{\alpha}_{t}}}(\boldsymbol{x}_{t}-\sqrt{1-\bar{\alpha}_{t}}\boldsymbol{\epsilon}_{\theta}(\boldsymbol{x}_{t}, t, \boldsymbol{x}_{c}, \boldsymbol{x}_{s})).
\end{align}

During training, at each step $t$, $\boldsymbol{x}_{0}$ is used to the following SCR module to compute the contrastive loss.

\section{Experiment Details}
\subsection{Categorization for Characters of Three Levels of Complexity}
\begin{figure}[h]
    \centering
    \subfloat[Easy]{
        \includegraphics[width=0.3\columnwidth]{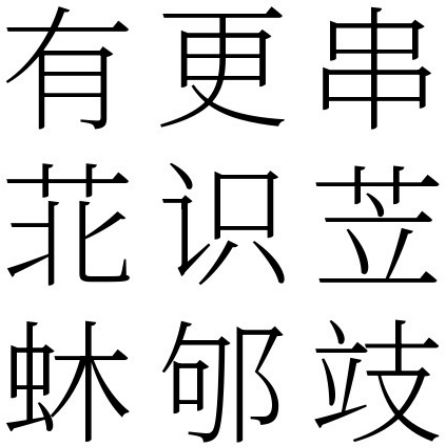}}
    \hfill
    \subfloat[Medium]{
        \includegraphics[width=0.3\columnwidth]{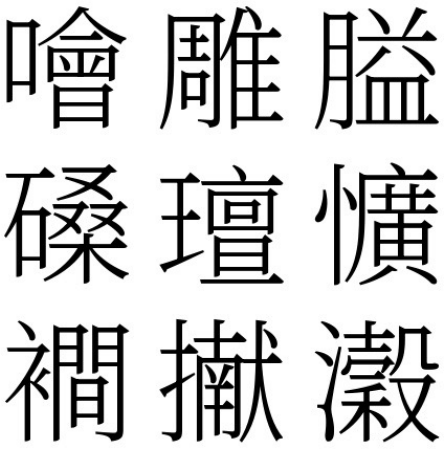}}
    \hfill
    \subfloat[Hard]{
        \includegraphics[width=0.3\columnwidth]{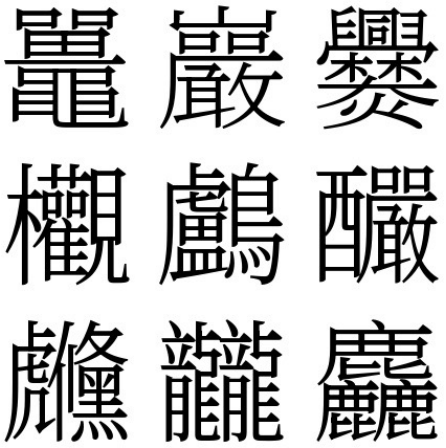}}
    \caption{Examples of three levels of complexity.}
    \label{fig:segment}
\end{figure}

To verify the effectiveness on characters of different complexity, we categorized the characters into three levels of complexity (easy, medium, and hard), according to their number of strokes. As illustrated in Table \ref{tab:complex_segment}, we categorized characters whose number of strokes is between 6 and 10 as characters of easy level, between 11 and 20 as medium level, and greater than 21 as hard level. Several categorization examples are shown in Figure \ref{fig:segment}. 

\begin{table}[h]
\centering
\resizebox{5cm}{!}{%
\begin{tabular}{@{}c|c@{}}
\toprule
\begin{tabular}[c]{@{}c@{}}complexity level\end{tabular} & stroke number $M$ \\ \midrule
Easy          & $6\le M \le 10$      \\
Medium        & $11\le M \le 20$      \\
Hard          & $M \ge 21$      \\ \bottomrule
\end{tabular}%
}
\caption{Categorization for three levels of complexity.}
\label{tab:complex_segment}
\end{table}

\subsection{Implementation Details}
Our training procedure adopts a coarse-to-fine two-phase strategy. And during phase 2, we employ a pre-trained SCR as a supervisor. In this section, we provide the pre-training details of SCR.
\subsubsection{Pre-training of SCR} We pre-train the Style Contrastive Refinement (SCR) module by AdamW optimizer, with $lr=1e-4$, 1000 warm-up steps, and linear learning rate schedule. The number of negative samples during pre-training is set as $48$ and the image size is set as $96$. The training set includes 400 fonts and 800 characters (the same as the training data in Chinese font generation of our experiments). SCR is supervised by the style contrastive loss $\mathcal{L}_{sc}^{SCR}$ as:
\begin{small}
\begin{align}
    \mathcal{L}_{sc}^{SCR} = -\sum_{l=0}^{N_{p}-1}log\frac{exp(v_{tar}^{l}\cdot v_{p}^{l}/\tau)}{exp(v_{tar}^{l}\cdot v_{p}^{l}/\tau)+\sum_{i=1}^{K}exp(v_{tar}^{l}\cdot v_{n_{i}}^{l}/\tau)},
\end{align}
\end{small}
where $v_{tar}$ dennotes the target image. $v_{p}$ and $v_{n}$ represent the positive sample (augmented target image) and negative sample (with different styles but the same character). The augmentation on positive images includes random cropping and random resizing. $K$ is the number of chosen negative samples and is set as $48$ during pre-training. During pre-training, $N_{p}$ is the number of the chosen VGG layer features and we choose the features $\boldsymbol{F}_{v}=\{f_{v}^{0}, f_{v}^{1}, f_{v}^{2}, f_{v}^{3}, f_{v}^{4}, f_{v}^{5}\}$ ($f_{v}^{i}$ is the ReLU output of i-th VGG convolution block). 

\subsection{More Ablation Studies}
\subsubsection{Influence of negative samples for $\mathcal{L}_{sc}$ in phase 2} We further discuss the influence of the numbers of negative samples for $\mathcal{L}_{sc}$, as shown in the Table \ref{tab:negative_samples}. The results of $K=16$ and $K=32$ are comparable, and we adopt the setting $K=16$ in all our experiments due to the reduction of its training time. 

\begin{table}[h]
\centering
\resizebox{7cm}{!}{%
\begin{tabular}{@{}c|cccc@{}}
\toprule
\begin{tabular}[c]{@{}c@{}}negative\\ samples $K$\end{tabular} & FID$\downarrow$            & SSIM$\uparrow$            & LPIPS$\downarrow$           & L1$\downarrow$              \\ \midrule
8        & 8.5900          & 0.4148          & 0.1501          & 0.3919          \\
16       & 8.5352          & \textbf{0.4206} & 0.1496          & \textbf{0.3870} \\
32       & \textbf{8.0692} & \underline{0.4174}          & \textbf{0.1487} & \underline{0.3897}          \\
48       & \underline{8.2454}          & 0.4172          & \underline{0.1495}          & 0.3899          \\ \bottomrule
\end{tabular}
}
\caption{Influence of the number of negative samples for $\mathcal{L}_{sc}$. The bold indicates the state-of-the-art and the underline indicates the second best.}
\label{tab:negative_samples}
\end{table}

\subsubsection{The influence of VGG layer features in SCR} We further discuss the influence of the VGG layer features $\boldsymbol{F}_{v}=\{f_{v}^{0}, f_{v}^{1}, ..., f_{v}^{N}\}$ in SCR during phase 2 ($f_{v}^{i}$ is the ReLU output of i-th VGG convolution block). As shown in Table \ref{tab:layer_features}, employing multi-scale VGG features can effectively boost the performance, and the setting $\boldsymbol{F}_{v}=\{f_{v}^{0}, f_{v}^{1}, f_{v}^{2}, f_{v}^{3}\}$ can obtain the best quality of our generation.

\begin{table}[h]
\centering
\resizebox{7cm}{!}{%
\begin{tabular}{@{}l|cccc@{}}
\toprule
\begin{tabular}[c]{@{}c@{}}layer\\ features $\boldsymbol{F}_{v}$\end{tabular} & FID$\downarrow$             & SSIM$\uparrow$ & LPIPS$\downarrow$           & L1$\downarrow$              \\ \midrule
$f_{v}^{3}$                                                        & 9.2220          & 0.4170    & 0.1527          & \underline{0.3890}          \\
$f_{v}^{2}, f_{v}^{3}$                                                      & \underline{8.2554}          & 0.4167    & \underline{0.1499}          & 0.3902          \\
$f_{v}^{1}, f_{v}^{2}, f_{v}^{3}$                                                    & \textbf{8.2173} & 0.4166    & 0.1505          & 0.3906          \\
$f_{v}^{0}, f_{v}^{1}, f_{v}^{2}, f_{v}^{3}$                                                  & 8.5352          & 0.4206    & \textbf{0.1496} & \textbf{0.3870} \\ \bottomrule
\end{tabular}%
}
\caption{Influence of VGG layer features $\boldsymbol{F}_{v}$ in phase 2.}
\label{tab:layer_features}
\end{table}

\subsubsection{Influence of guidance scales} We further discuss the influence of guidance scales $s$ during sampling. As shown in Table \ref{tab:guidance_scale}, the setting $s=7.5$ 
achieves the best performance.

\begin{table}[h]
\centering
\resizebox{6.5cm}{!}{%
\begin{tabular}{@{}c|cccc@{}}
\toprule
\begin{tabular}[c]{@{}c@{}}guidance\\ scales $s$\end{tabular} & FID$\downarrow$             & SSIM$\uparrow$            & LPIPS$\downarrow$           & L1$\downarrow$              \\ \midrule
1                                                         & \textbf{7.1447} & 0.3826          & 0.1696          & 0.4223          \\
3.5                                                       & 8.5504          & 0.4137          & 0.1504          & 0.3929          \\
5.5                                                       & \underline{8.4842}          & 0.4188          & \textbf{0.1496} & 0.3885          \\
7.5                                                       & 8.5352          & \textbf{0.4206} & \textbf{0.1496} & \textbf{0.3870} \\
9.5                                                       & 8.8995          & 0.4198          & 0.1503          & \underline{0.3873}          \\
11.5                                                      & 9.6069          & \underline{0.4201}          & 0.1514          & \underline{0.3873}          \\
15                                                        & 12.2369         & 0.4194          & 0.1532          & 0.3880          \\
20                                                        & 18.2550         & 0.4175          & 0.1581          & 0.3899          \\
30                                                        & 45.3899         & 0.4087          & 0.1790          & 0.3964          \\ \bottomrule
\end{tabular}%
}
\caption{Influence of guidance scales $s$.}
\label{tab:guidance_scale}
\end{table}

\subsection{Limitations}
Though we adopt the efficient sampler DPM-Solver++ \cite{lu2022dpm}, our method still needs to generate the sample in a few steps as most diffusion-based generation methods (the speed is slower than GAN-based methods). 

\subsection{More Visualization of the Results}
\begin{figure*}[t]
    \centering
    \subfloat[Characters of easy level of complexity]{
        \includegraphics[width=\textwidth]{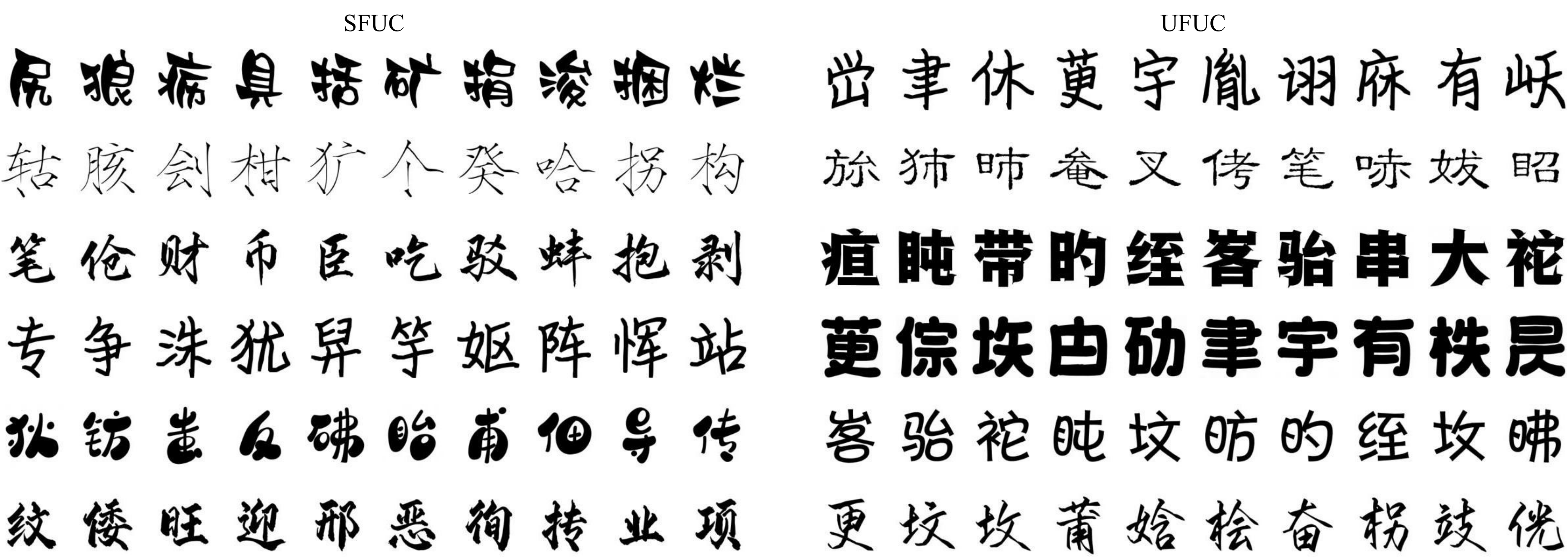}}
    \vspace{0.65cm}
    \subfloat[Characters of medium level of complexity]{
        \includegraphics[width=\textwidth]{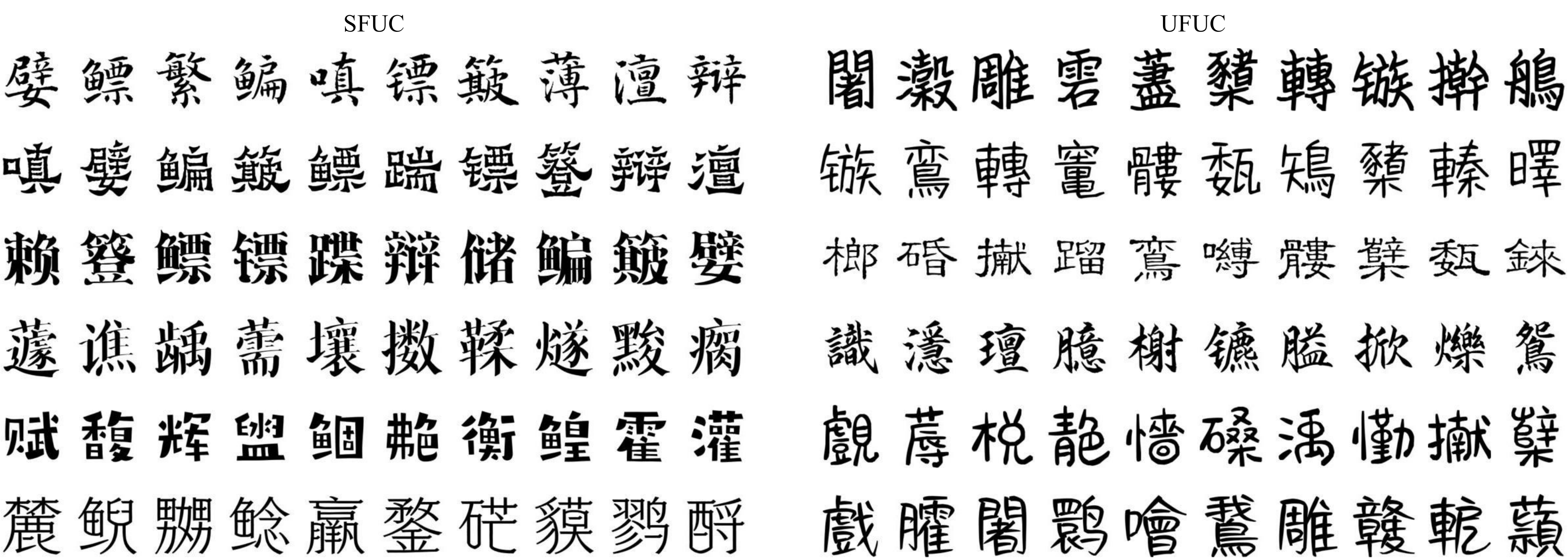}}
    \vspace{0.65cm}
    \subfloat[Characters of hard level of complexity]{
        \includegraphics[width=\textwidth]{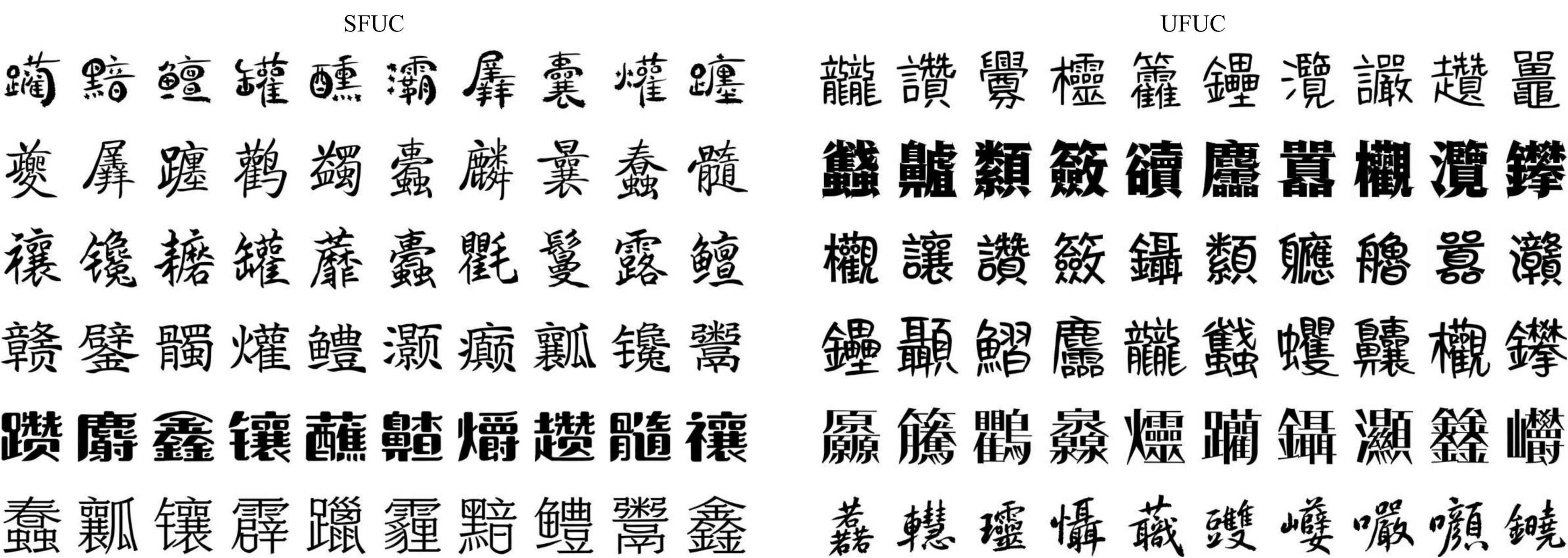}}    
    \caption{Visualization of the results by FontDiffuser.}
    \label{fig:more_results}
\end{figure*}

\begin{figure*}[t]
    \centering
    \includegraphics[width=\textwidth]{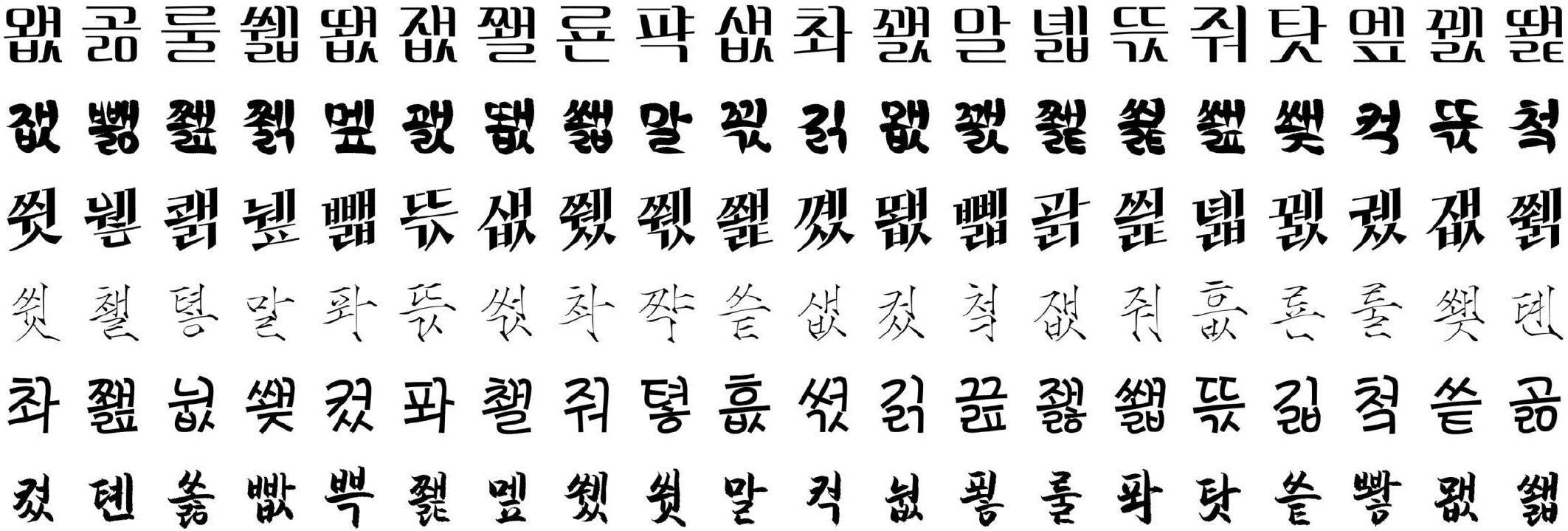}
    \caption{Visualization of cross-lingual generation (Chinese to Korean). }
    \label{fig:more_kor}
\end{figure*}

In this section, we provide more visualization of the results generated by FontDiffuser. As shown in Figure \ref{fig:more_results}, the Chinese font generation results include the generated characters of three levels of complexity (easy, medium, and hard) on Seen Font Unseen Character (SFUC) and Unseen Font Unseen Character (UFUC). Additionally, we also provide more visualization of the cross-lingual generation (Chinese to Korean) by FontDiffuser, as shown in Figure \ref{fig:more_kor}.

\end{document}